\documentclass{article}
\usepackage{arxiv}
\usepackage{hyperref}
\usepackage{url}
\usepackage{booktabs}
\usepackage{amsfonts}
\usepackage{nicefrac}
\usepackage{microtype}
\usepackage{graphicx}
\usepackage{natbib}
\usepackage{doi}
\usepackage{tabularx}
\usepackage{amsmath} 
\usepackage{multirow}

\title{FAD-SA-GRU: Enhancing Hate Speech Detection in Algerian Dialect through Feature-Augmented Self-Attention GRU Networks}

\author{ {(1)Sara YAKOUBI, (1)Ikram KHALFALLAH, (2)Kenza KHELKHAL, (2)Dihia LANASRI}\\
	(1)USTHB, (2)ATM Mobilis \\
	Algiers, Algeria\\	
	\texttt{yakoubi.sara21@gmail.com, ikramkhalf736@gmail.com} \\
\texttt{khelkhalkenza88@gmail.com, dihia.lanasri@gmail.com}
    \\    
}

\renewcommand{\shorttitle}{FAD-SA-GRU: Hate Speech Detection in Algerian Dialect}

\hypersetup{
pdftitle={Hate Speech Detection in Algerian Dialect},
pdfsubject={NLP, DIALECT},
pdfauthor={S.YAKOUBI and al},
pdfkeywords={Natural Language Processing, Hate Speech Detection, Algerian Dialect, GRU, Deep Learning, FAD-SA-GRU},
}

\begin{document}
\maketitle

\begin{abstract}
The widespread adoption of social media platforms has transformed online communication by enabling users to exchange information and opinions instantly. However, these platforms have also facilitated the rapid dissemination of abusive and hateful content, posing significant social, psychological, and ethical challenges. Hate speech can incite discrimination, harassment, and violence against individuals or communities based on attributes such as ethnicity, religion, gender, nationality, or political affiliation. Consequently, the automatic detection of hate speech has become a major research topic in natural language processing (NLP), attracting considerable attention from both academia and industry as an essential component of content moderation systems.

This paper investigates the automatic detection of hate speech in Algerian Arabic dialect (Darija) on social media. Hate speech detection in Algerian Darija remains a challenging task due to the dialect's linguistic diversity, characterized by the coexistence of Arabic, French, and Arabizi (Arabic written using the Latin alphabet). To address these challenges, we conduct a comprehensive comparative study of four categories of text classification approaches: (1) traditional machine learning models using TF--IDF feature representations, (2) deep learning models based on recurrent neural networks, (3) Transformer-based language models, including DziriBERT and multilingual BERT, and (4) a novel hybrid architecture, \textbf{FAD-SA-GRU}, which combines semantic representations from DZ FastText, DZ AraVec, and DziriBERT through a multi-embedding fusion strategy, followed by a self-attention-enhanced GRU encoder.

Experiments are conducted on a manually annotated dataset of Algerian Darija social media comments labeled for binary hate speech classification. The proposed FAD-SA-GRU model consistently outperforms all baseline approaches, achieving an accuracy of 93.2\%, precision of 93.4\%, recall of 91.0\%, F1-score of 92.1\%, and ROC-AUC of 97.0\%. The experimental results demonstrate the effectiveness of combining complementary embedding representations with attention-based sequence modeling for robust hate speech detection in low-resource dialectal Arabic.

\end{abstract}


\section{Introduction}
The rapid expansion of social media has profoundly transformed the way people communicate, share information, and express opinions. Platforms such as Facebook, X (formerly Twitter), Instagram, and YouTube have become indispensable channels for public discourse, allowing millions of users to interact in real time regardless of geographical boundaries. While these platforms promote connectivity, collaboration, and freedom of expression, they have also facilitated the proliferation of harmful online content, including cyberbullying, misinformation, abusive language, and hate speech. The sheer volume of user-generated content makes manual moderation increasingly impractical, motivating the development of automated solutions capable of detecting and filtering harmful content efficiently. Consequently, hate speech detection has emerged as a major research topic in Natural Language Processing (NLP), with applications in online content moderation, public safety, and the protection of vulnerable communities.

Hate speech is generally understood as any form of communication that attacks, threatens, or discriminates against an individual or a group based on protected characteristics such as ethnicity, nationality, religion, race, gender, or political affiliation. Beyond its direct impact on targeted individuals, hate speech contributes to social polarization, fosters discrimination, and may incite real-world violence. Governments, social media companies, and researchers have therefore invested considerable effort in developing computational methods capable of identifying hateful content automatically. Recent advances in machine learning, deep learning, and Transformer-based language models have significantly improved hate speech detection performance in high-resource languages, particularly English. Nevertheless, the effectiveness of these approaches remains limited for low-resource languages and dialects due to the scarcity of annotated datasets, linguistic resources, and domain-specific pretrained models.

Among Arabic dialects, Algerian Arabic (Darija) presents particularly challenging characteristics for automatic language processing. Unlike Modern Standard Arabic, Darija has no standardized orthography, resulting in multiple spellings for the same lexical item. Moreover, online users frequently alternate between Arabic and French through code-switching and commonly employ \emph{Arabizi}, where Arabic words are written using the Latin alphabet and numerals (e.g., ``3'', ``7'', and ``9''). These linguistic phenomena produce highly heterogeneous textual data characterized by lexical variation, inconsistent spelling, and mixed writing systems. Such characteristics substantially complicate feature extraction, semantic representation, and text classification, making hate speech detection in Algerian Darija considerably more difficult than in standardized languages.

Although the NLP community has devoted increasing attention to Arabic language processing, research on Algerian Darija remains relatively limited. Most existing studies focus either on Modern Standard Arabic or on other Arabic dialects for which larger corpora and pretrained language models are available \cite{}. 
Furthermore, many previous works evaluate only a single family of classification methods \cite{}, 
making it difficult to assess the relative strengths and limitations of traditional machine learning algorithms, deep neural architectures, and Transformer-based models under comparable experimental conditions. In addition, relatively few studies investigate the integration of multiple complementary semantic representations to improve hate speech detection in Algerian dialectal Arabic \cite{lanasri2023hate}.

This work addresses these limitations by investigating automatic hate speech detection in Algerian Darija using comments collected from Facebook related to telecom. These public comments reflecting diverse opinions, customer experiences, and discussions. While many interactions are constructive, a non-negligible proportion contains abusive or hateful language directed toward the company, its employees, or other users. The automatic identification of such content is therefore essential for assisting moderators, protecting the company's public image, and improving the efficiency of content management systems.

To tackle this problem, we conduct a comprehensive comparative evaluation of four families of hate speech detection approaches. First, we investigate traditional machine learning algorithms based on TF--IDF representations. Second, we evaluate recurrent neural network architectures capable of modeling sequential textual information. Third, we assess the effectiveness of state-of-the-art Transformer models, including multilingual BERT and DziriBERT, which have demonstrated strong contextual language understanding capabilities. Building upon the insights obtained from these comparative experiments, we propose a novel hybrid architecture, \textbf{FAD-SA-GRU}, which combines complementary semantic representations obtained from DZ FastText, DZ AraVec, and DziriBERT through a multi-embedding fusion strategy. A self-attention mechanism is subsequently applied over the GRU encoder outputs to emphasize the most informative contextual features before the final classification layer.

The principal contributions of this work can be summarized as follows:
\begin{itemize}
    \item We constructed and annotated a dataset of 22,193 comment in Algerian Darija social media comments for binary hate speech detection.    
    \item We performed a comprehensive comparative evaluation of traditional machine learning methods, recurrent deep learning architectures, and Transformer-based language models for hate speech detection in Algerian Darija.    
    \item We proposed \textbf{FAD-SA-GRU}, a novel hybrid architecture that combines DZ FastText, DZ AraVec, and DziriBERT embeddings using a multi-embedding fusion strategy together with a self-attention-enhanced GRU encoder.  
\end{itemize}

The remainder of this paper is organized as follows. Section 02 reviews the related literature. Section 03 presents the proposed methodology, dataset, preprocessing pipeline, and model architectures. Section 04 reports the experimental results, while Section 05 discusses their implications. Finally, Section 06 concludes the paper and outlines future research directions.

\section{Related Work}
\label{sec:related}
Hate speech generally refers to any form of communication that attacks, demeans, threatens, or incites hostility against an individual or a group based on protected characteristics such as ethnicity, race, religion, nationality, gender, language, etc. \cite{perez2023assessing}. The automatic identification of such content has become a major research topic in Natural Language Processing (NLP), driven by the exponential growth of user-generated content on social media and the increasing need for effective content moderation systems. Over the past decade, numerous computational approaches have been proposed, ranging from traditional machine learning algorithms relying on handcrafted textual features to deep neural networks and, more recently, Transformer-based language models.

Early studies primarily employed conventional machine learning techniques such as Support Vector Machines (SVM), Naïve Bayes (NB), Logistic Regression (LR), and Random Forests (RF), using feature representations based on bag-of-words, TF--IDF, and lexical or syntactic characteristics. Although these methods achieved encouraging results on relatively simple datasets, their performance was often limited by their inability to capture contextual semantics and long-range linguistic dependencies. The emergence of deep learning significantly improved hate speech detection by enabling models to automatically learn semantic representations from text through architectures such as Convolutional Neural Networks (CNNs), Long Short-Term Memory (LSTM) networks, and Gated Recurrent Units (GRUs). More recently, Transformer-based models, including BERT and its multilingual variants, have established the state of the art across numerous text classification tasks by leveraging contextualized language representations learned through large-scale pretraining.

In what follows, we will review the most relevant contributions in the literature, beginning with studies conducted on high-resource languages before discussing research dedicated to Arabic and, more specifically, Algerian Arabic (Darija).

\subsection{Hate Speech Detection in High-Resource Languages}
Early approaches primarily relied on conventional machine learning algorithms using lexical and statistical representations such as Bag-of-Words and TF--IDF. Although computationally efficient, these methods generally struggle to capture contextual semantics and long-range linguistic dependencies. To overcome these limitations, \cite{khezzar2023} introduced \textit{arHateDetector}, a framework for hate speech detection in Arabic supporting both Modern Standard Arabic (MSA) and several Arabic dialects. The authors constructed \textit{arHateDataset}, a balanced corpus of 34,107 annotated tweets, and applied an extensive preprocessing pipeline including text normalization, stop-word removal, and lemmatization using FARASA. Their experimental evaluation compared nine traditional machine learning classifiers, including LinearSVC, Logistic Regression, and Random Forest, against deep learning models such as CNN and the pretrained Transformer AraBERT. The results demonstrated the superiority of contextual language models, with AraBERT achieving an accuracy of 93\%, outperforming LinearSVC (89\%) and CNN (88\%).

The effectiveness of recurrent neural architectures for Arabic hate speech detection was further investigated by \cite{bouchal2023}, who addressed the identification of hateful and offensive content related to the COVID-19 pandemic on Twitter. Using the \textit{Ar-hatespeech-off} dataset containing 8,662 Arabic messages, the authors mitigated class imbalance through undersampling before training a Bidirectional Long Short-Term Memory (Bi-LSTM) network with pretrained word embeddings. Their approach achieved a precision of 96.35\% and an F1-score of 85.82\%, confirming the ability of recurrent neural networks to capture sequential contextual information more effectively than conventional feature-based classifiers.

The emergence of Transformer architectures has substantially improved hate speech detection across multiple languages. In the context of French, \cite{chared2025} conducted a comprehensive comparative study using a corpus composed of authentic tweets, synthetic sentences, and AI-generated examples. Their evaluation included traditional machine learning algorithms (SVM, Logistic Regression, Random Forest, and Naive Bayes), recurrent neural networks (LSTM, Bi-LSTM, and GRU), and several Transformer models, including CamemBERT, DistilCamemBERT, and DeHateBERT. While classical machine learning models achieved competitive F1-scores around 0.78, recurrent neural architectures produced comparatively lower performance. Transformer models consistently obtained the best results, with DistilCamemBERT achieving an F1-score of 0.80, followed closely by DeHateBERT with a recall of 0.82. The authors further incorporated Local Interpretable Model-Agnostic Explanations (LIME) to improve model interpretability and demonstrate the influence of contextual information on classification decisions.

Similarly, \cite{saleh2023} investigated hate speech detection in English by comparing specialized word embeddings with pretrained Transformer models. They introduced \textit{HSW2V} (Hate Speech Word2Vec), a domain-specific embedding model trained on more than one million sentences collected from multiple public hate speech datasets and Twitter posts. The proposed HSW2V embeddings were combined with a BiLSTM classifier and compared against fine-tuned BERT Base and BERT Large models. Experimental results showed that the HSW2V--BiLSTM architecture achieved an F1-score of 93\%, outperforming conventional Word2Vec and GloVe embeddings despite being trained on a smaller corpus. Nevertheless, BERT Large obtained the best overall performance with an F1-score of 96\%, highlighting the advantages of large-scale contextual pretraining. The study also employed LIME to analyze prediction explanations and demonstrated the robustness of HSW2V for handling misspelled and non-standard words.

Hybrid deep learning architectures have also demonstrated promising performance. \cite{kumar2024} proposed a BiLSTM--CNN architecture for multilingual hate speech detection across several social media platforms. Their model combines GloVe word embeddings with a Bidirectional LSTM encoder to capture contextual dependencies, followed by a convolutional neural network for extracting discriminative local features. Using a heterogeneous corpus obtained by merging ten publicly available hate speech datasets, the proposed architecture achieved an overall accuracy of 95.75\%, with precision, recall, and F1-score all close to 95\%. Particularly strong performance was reported on Facebook, YouTube, and Gab datasets, demonstrating the effectiveness of combining sequential and convolutional feature extraction mechanisms.

Recent work has also explored the use of Large Language Models (LLMs) and Small Language Models (SLMs) for hate speech detection. \cite{delaval2026} introduced \textit{TOXIFRENCH}, a corpus specifically designed for French toxic language identification, and compared three categories of models: LLMs such as GPT-4o, BERT-based classifiers including CamemBERT, and lightweight SLMs based on Qwen3-4B. To improve reasoning capabilities, the authors incorporated a Chain-of-Thought (CoT) prompting strategy together with a Dynamic Weighted Loss function that prioritized final classification decisions during training. Their optimized Qwen3-4B model achieved a precision of 86.5\%, demonstrating that carefully specialized lightweight language models can provide competitive performance while requiring substantially fewer computational resources than very large models.

Overall, the literature on high-resource languages reveals a clear progression from traditional feature-based machine learning methods toward deep neural architectures and Transformer-based language models. While classical classifiers remain attractive because of their simplicity and computational efficiency, contextual language models consistently achieve superior performance by capturing richer semantic and contextual information. More recent studies further suggest that combining complementary neural architectures, specialized embeddings, and pretrained language models can yield additional performance improvements. These findings motivate the exploration of hybrid architectures for low-resource dialects such as Algerian Darija, where linguistic variability and limited annotated resources present additional challenges.

\subsection{Hate Speech Detection in Algerian Darija}
Compared with high-resource languages, research on hate speech detection in Algerian Arabic (Darija) remains relatively limited. The linguistic complexity of Darija, characterized by the coexistence of Modern Standard Arabic, dialectal Arabic, French, Amazigh borrowings, and Arabizi, together with the absence of standardized orthographic conventions, presents significant challenges for automatic language processing. Furthermore, the scarcity of publicly available annotated corpora has constrained the development and evaluation of robust hate speech detection systems. Nevertheless, recent studies have made important contributions by introducing dedicated datasets and exploring various machine learning and deep learning approaches.

One of the earliest contributions was presented by \cite{guellil2021}, who investigated the automatic detection of sexist speech in Algerian online communities. The authors constructed a corpus of approximately 5,000 YouTube comments written in Arabic, Algerian Darija, French, and English, which were manually annotated by three native speakers into sexist and non-sexist categories. Several traditional machine learning algorithms and deep learning models, including Logistic Regression, SVM, CNN, LSTM, and Bi-LSTM, were evaluated using Word2Vec and FastText embeddings. Among the evaluated architectures, the CNN model achieved the best performance with an F1-score of 86\%, demonstrating the effectiveness of convolutional neural networks for identifying implicit lexical patterns in multilingual Algerian social media content.

To further enrich the available linguistic resources, \cite{boucherit2022} introduced \textit{DziriOFN}, a corpus comprising more than 8,700 Algerian dialect texts manually annotated into three categories: normal, abusive, and offensive. The authors compared conventional machine learning classifiers, including SVM, Multinomial Naïve Bayes, and Gaussian Naïve Bayes, with several deep learning architectures such as CNN, BiLSTM, and FastText. Their experiments revealed that classical statistical methods remained highly competitive, with SVM and Multinomial Naïve Bayes achieving accuracies of 74\% and 75\%, respectively, outperforming the evaluated deep learning models. These findings highlight the difficulty of applying deep neural networks to relatively small Algerian dialect datasets.

A more comprehensive study was conducted by \cite{lanasri2023hate}, who proposed the \textit{DzaraShield} framework for hate speech detection in Algerian Darija. Their corpus, consisting of more than 13,500 manually annotated comments collected from Facebook, Twitter, and YouTube, was used to evaluate both traditional machine learning algorithms and modern deep learning architectures. The proposed framework achieved a precision, recall, and F1-score of 87\%, demonstrating the potential of deep learning techniques when trained on larger and more diverse corpora.

Focusing on a broader taxonomy of toxic language, \cite{mazari2023} developed a multilingual corpus of 14,150 Facebook, Twitter, and YouTube comments annotated according to three categories: hate speech, offensive language, and cyberbullying. Following linguistic preprocessing, the authors compared conventional classifiers such as SVM and Naïve Bayes with recurrent neural architectures including LSTM and Bidirectional GRU (Bi-GRU). Their results showed that the Bi-GRU model produced the best overall performance, achieving a precision of 73.6\% and an F1-score of 75.8\%, confirming the benefits of bidirectional recurrent architectures for modeling contextual dependencies in dialectal Arabic.

More recently, Transformer-based language models have begun to demonstrate promising results for Algerian dialect processing. \cite{abdedaiem2024} introduced \textit{FASSILA}, the first large-scale corpus specifically designed for fake news detection and sentiment analysis in Algerian Darija. The dataset contains 10,087 manually annotated sentences collected from YouTube, Facebook, and previously published corpora, spanning seven thematic domains. The authors evaluated several machine learning algorithms together with pretrained Transformer models, including AraBERTv02, MARBERTv2, and DziriBERT. Their experiments showed that MARBERTv2 achieved the highest performance for both fake news detection (Accuracy 78.98\%, F1-score 77.39\%) and sentiment analysis (Accuracy 83.92\%, F1-score 80.35\%), emphasizing the importance of pretrained language models specifically adapted to Maghrebi dialects. Beyond the reported classification results, the study also highlighted the critical role of high-quality annotated corpora in advancing NLP research for Algerian Darija.

\subsection{Discussion and Research Gap}
The literature reviewed in the previous subsections demonstrates the significant evolution of hate speech detection methods, from traditional machine learning algorithms to deep learning and Transformer-based architectures. While contextual language models have substantially improved performance for high-resource languages, research on Algerian Darija remains comparatively limited due to the scarcity of annotated corpora and the linguistic complexity of the dialect, including code-switching, Arabizi, and the absence of standardized orthography.

Although recent studies have introduced valuable datasets and proposed effective classification models, several research gaps remain. Most existing works evaluate only a limited family of models, making comprehensive comparisons across machine learning, deep learning, and Transformer-based approaches difficult. Furthermore, the potential of combining complementary static and contextual embedding representations has received little attention, despite their ability to capture different semantic characteristics of dialectal text. Finally, few studies address the practical deployment of hate speech detection systems through reusable services suitable for real-world moderation scenarios.

To bridge these gaps, this work provides a unified comparative evaluation of traditional machine learning, recurrent neural networks, and pretrained Transformer models on the same Algerian Darija corpus. Building upon this analysis, we propose the \textbf{FAD-SA-GRU} architecture, which combines DZ FastText, DZ AraVec, and DziriBERT embeddings through a multi-embedding fusion strategy enhanced by self-attention. 

\section{Our Proposed Framework}
\label{sec:method}
This section presents the methodology adopted for automatic hate speech detection in Algerian Darija. The proposed framework comprises four successive stages: data collection and annotation, text preprocessing, model development, and performance evaluation. A unified experimental protocol is adopted to ensure a fair comparison among all investigated approaches, from traditional machine learning algorithms to deep learning and Transformer-based models. Building upon this comparative study, we introduce a novel hybrid architecture, \textbf{FAD-SA-GRU}, specifically designed to exploit complementary semantic representations of Algerian Darija.

\begin{figure}[h]
    \centering
    \includegraphics[width=\linewidth]{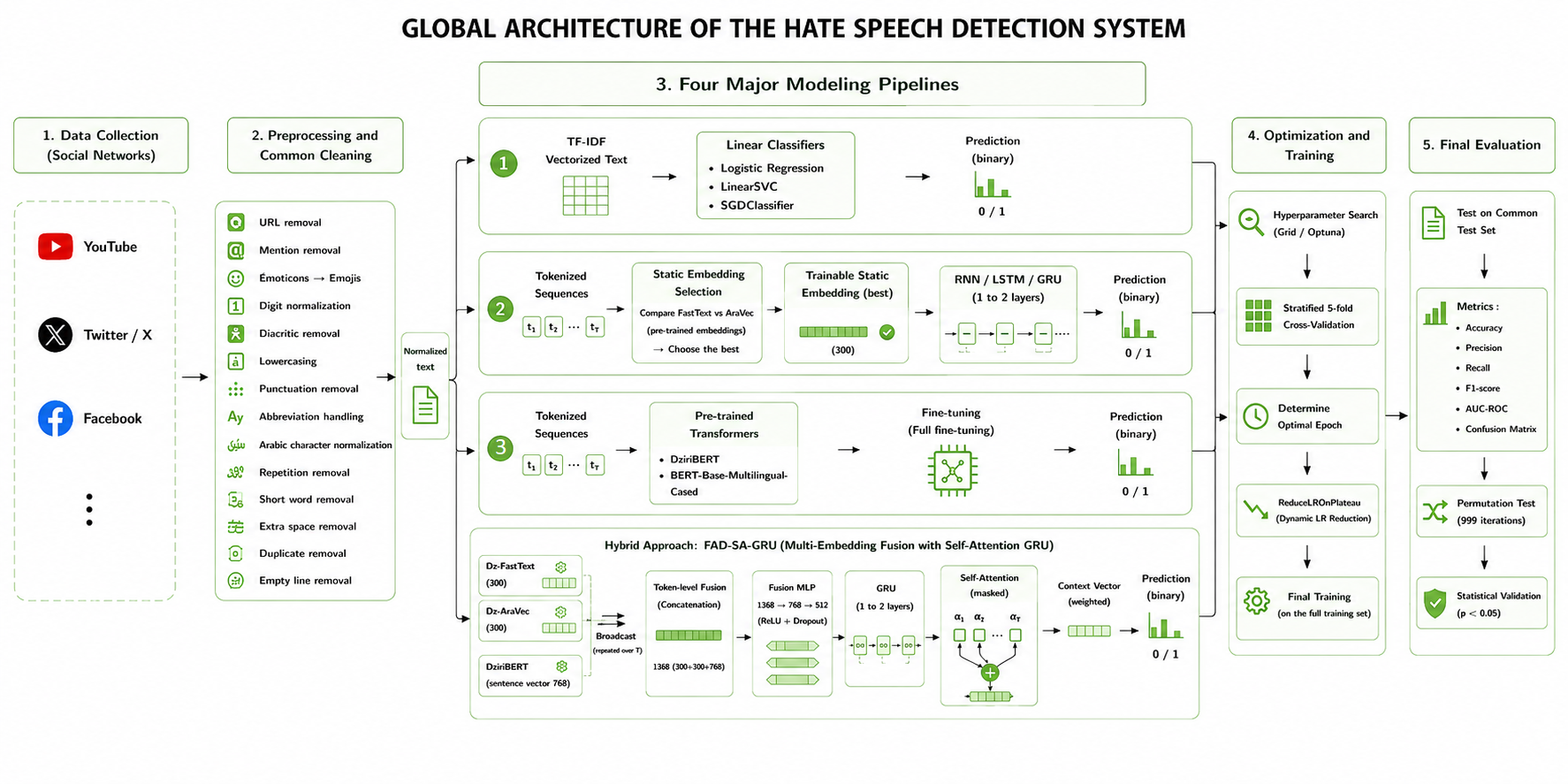}
    \caption{Overall methodology of the proposed hate speech detection framework.}
    \label{fig:arch-general}
\end{figure}

Figure~\ref{fig:arch-general} illustrates the overall architecture of the proposed framework. Social media comments are first collected and manually annotated to construct a binary hate speech corpus. The raw comments are then processed through a common preprocessing pipeline that normalizes Arabic and Arabizi text, removes noisy elements, and standardizes lexical representations. This preprocessing stage is shared by all subsequent models, ensuring that differences in performance are attributable to the classification architectures rather than inconsistencies in data preparation.

After preprocessing, the normalized corpus is used to train and evaluate four complementary families of hate speech detection models:

\begin{enumerate}
    \item \textbf{Traditional machine learning}, where comments are transformed into TF--IDF feature vectors and classified using linear statistical models.

    \item \textbf{Recurrent deep learning models}, in which tokenized sequences are represented through word embeddings models (FastText and AraVec) before being processed by recurrent neural networks, including RNN, LSTM, and GRU architectures.

    \item \textbf{Pretrained Transformer models}, namely DziriBERT and BERT-Base-Multilingual-Cased (mBERT), which are fine-tuned directly on the hate speech classification task.

    \item \textbf{The proposed FAD-SA-GRU architecture}, which combines static semantic representations from DZ FastText and DZ AraVec with contextual representations generated by DziriBERT through a multi-embedding fusion strategy. The fused representations are projected into a shared latent space, encoded by a Gated Recurrent Unit (GRU), and refined using a self-attention mechanism before the final binary classification layer.
\end{enumerate}

\subsection{Data Collection and Annotation}
The experimental dataset used in this study was constructed from user-generated comments collected from three widely used social media platforms: \textbf{Facebook}, \textbf{Twitter (X)}, and \textbf{YouTube}. These platforms were selected because they represent the primary online communication channels in Algeria and contain a large volume of comments written in Algerian Darija. Unlike formal written Arabic, social media comments frequently exhibit informal language, code-switching between Arabic and French, and the use of Arabizi, making them particularly suitable for evaluating hate speech detection systems under realistic conditions.

After collection, all comments were manually reviewed and annotated for hate speech detection. A binary annotation scheme was adopted, where comments were assigned to one of two classes: \textbf{non-hateful} (label 0) or \textbf{hateful} (label 1). The hateful class includes both explicit hate speech, characterized by direct insults, threats, or offensive expressions, and implicit hate speech, where hostility is conveyed through indirect language, sarcasm, or culturally dependent expressions. This binary formulation was chosen to align with the objectives of automated content moderation systems, which primarily require distinguishing harmful content from acceptable user interactions.

The resulting corpus contains a total of \textbf{22,193} annotated comments. The dataset comprises \textbf{12,472} non-hateful comments (56.2\%) and \textbf{9,721} hateful comments (43.8\%). Although the two classes are not perfectly balanced, the observed distribution does not exhibit severe imbalance and therefore does not require artificial resampling or data augmentation techniques. Instead, the original class proportions were preserved to better reflect the natural distribution of hate speech encountered on social media platforms.

The combination of multiple social media sources and the inclusion of comments written in both Arabic script and Arabizi provide a linguistically diverse corpus that captures the variability of real-world Algerian online communication. This diversity enables a comprehensive evaluation of the proposed hate speech detection models under realistic and challenging conditions.

\subsection{Common Preprocessing Pipeline}
Given the heterogeneous nature of Algerian Darija on social media, a unified preprocessing pipeline is applied to all collected comments before model training. The objective is to reduce lexical variability, remove irrelevant information, and normalize both Arabic and Arabizi texts while preserving the semantic content required for hate speech detection. Since all classification models share the same preprocessing stage, any observed performance differences can be attributed to the learning architectures rather than inconsistencies in data preparation.

The preprocessing pipeline consists of the following operations:

\begin{enumerate}

\item \textbf{Removal of URLs and user mentions.}
Hyperlinks and user mentions (e.g., \texttt{@username}) are removed because they do not contribute to the semantic interpretation of hate speech and unnecessarily increase vocabulary sparsity.

\item \textbf{Emoji normalization.}
Textual emoticons (e.g., \texttt{:)} or \texttt{:(}) are converted into their corresponding Unicode emojis using a manually constructed mapping dictionary, providing a consistent representation of users' emotional expressions.

\item \textbf{Digit normalization.}
Arabic-Indic numerals are converted into Western digits (0--9) to ensure a uniform numerical representation throughout the corpus.

\item \textbf{Arabic text normalization.}
Arabic diacritics are removed, and characters with multiple orthographic variants are mapped to a single canonical form. This step reduces vocabulary fragmentation while preserving lexical meaning.

\item \textbf{Lowercasing.}
All Latin characters, including French words and Arabizi text, are converted to lowercase to eliminate case-related lexical variations.

\item \textbf{Punctuation removal.}
Arabic and Latin punctuation marks are removed using a predefined multilingual punctuation list, reducing textual noise without affecting semantic content.

\item \textbf{Abbreviation expansion.}
Frequently used abbreviations and informal expressions in Arabizi, French, and English are expanded using a manually compiled multilingual dictionary containing more than 130 entries. For example, \texttt{mrc} is replaced by \textit{merci}, while \texttt{wsh} is normalized to \textit{wach}.

\item \textbf{Repeated-character normalization.}
Character elongations commonly used to express emphasis (e.g., \texttt{tooooop}) are reduced to a maximum of two consecutive identical characters, preserving readability while limiting vocabulary sparsity.

\item \textbf{Short-token filtering.}
Isolated tokens shorter than two characters are discarded, as they generally correspond to noise or carry little semantic information.

\item \textbf{Corpus cleaning.}
Finally, redundant whitespace, duplicate comments, and empty records are removed to obtain a clean and consistent corpus suitable for model training.
\end{enumerate}

The proposed preprocessing pipeline considerably reduces lexical variability while preserving the linguistic characteristics of Algerian Darija. In particular, the normalization of Arabizi and Arabic orthographic variants enables more consistent word representations across all embedding methods, thereby improving the effectiveness of both traditional machine learning algorithms and neural language models.

\subsection{Model Development}
\subsubsection{Approach 01: Classical Machine Learning Models}
For the classical machine learning approaches, the common preprocessing pipeline is complemented by a stop-word removal step using the Arabic and French stop-word lists provided by NLTK, together with a manually constructed list of frequent Algerian Darija stop words. The resulting documents are represented using the TF--IDF scheme.

Three linear classifiers are evaluated: Logistic Regression, Linear Support Vector Classification (LinearSVC), and SGDClassifier. Hyperparameters for each model were optimized using GridSearchCV with 5-fold cross-validation.

\subsubsection{Approach 02: Deep Learning with Recurrent Neural Networks}
\label{deeplearning-training}
Three recurrent neural architectures are investigated for hate speech detection: the standard Recurrent Neural Network (RNN), the Long Short-Term Memory (LSTM) network, and the Gated Recurrent Unit (GRU). All three models share the same overall architecture and differ only in the recurrent unit employed to model sequential dependencies within the input text.

Each comment is first tokenized and padded to a fixed length of 100 tokens. Two pretrained 300-dimensional word embedding models are evaluated: \textbf{AraVec} and \textbf{FastText}. In the first stage, both embeddings remain frozen in order to assess their intrinsic representational quality. The best-performing embedding is subsequently fine-tuned on the training corpus, producing a domain-adapted representation referred to as \textbf{Dz FastText}. Across all recurrent architectures, FastText consistently achieved superior performance and was therefore selected for fine-tuning.

Hyperparameter optimization is performed using Optuna in conjunction with stratified five-fold cross-validation. 

Table~\ref{tab:hyperparams-dl} summarizes the optimal hyperparameter configuration obtained for each recurrent architecture and embedding combination.

\begin{table}[h!]
\centering
\caption{Essential hyperparameters found for the deep learning models.}
\label{tab:hyperparams-dl}
\resizebox{\textwidth}{!}{%
\begin{tabular}{lcccccccc}
\toprule
Model & Hidden dim & Layers & Dropout (pre) & Dropout & Dropout (post) & Weight decay & Learning rate & Optimal epoch \\
\midrule
FastText + RNN       & 64  & 1 & 0.2 & 0.4 & 0.5 & 0.00026 & 0.00048 & 10 \\
AraVec + RNN         & 96  & 1 & 0.2 & 0.6 & 0.3 & 0.00026 & 0.00081 & 7  \\
Dz FastText + RNN    & 32  & 2 & 0.5 & 0.6 & 0.3 & 0.00011 & 0.00037 & 3  \\
FastText + GRU       & 96  & 1 & 0.2 & 0.3 & 0.5 & 0.00026 & 0.00037 & 10 \\
AraVec + GRU         & 80  & 1 & 0.3 & 0.4 & 0.2 & 0.00024 & 0.00375 & 10 \\
Dz FastText + GRU    & 128 & 2 & 0.4 & 0.7 & 0.2 & 0.00012 & 0.00012 & 6  \\
FastText + LSTM      & 96  & 2 & 0.3 & 0.4 & 0.3 & 0.00034 & 0.00040 & 10 \\
AraVec + LSTM        & 48  & 1 & 0.2 & 0.4 & 0.4 & 0.00038 & 0.00037 & 10 \\
Dz FastText + LSTM   & 96  & 2 & 0.5 & 0.6 & 0.2 & 0.00011 & 0.00022 & 4  \\
\bottomrule
\end{tabular}%
}
\end{table}

\subsubsection{Approach 03: Pre-trained Transformer Models}
Transformer-based language models have recently established the state of the art for numerous Natural Language Processing (NLP) tasks owing to their ability to learn rich contextual representations through self-attention mechanisms. In this work, two BERT-based models are investigated for hate speech detection in Algerian Darija: \textbf{DziriBERT} \cite{abdaoui2021dziribert}, specifically pre-trained for the Algerian dialect, and the multilingual \textbf{BERT-Base-Multilingual-Cased (mBERT)} \footnote{\url{https://github.com/google-research/bert/blob/master/multilingual.md}}.

For each model, two experimental settings are considered. First, the pretrained model is used as a fixed feature extractor, where the encoder weights remain frozen and only the classification layer is trained. This setting evaluates the transferability of the pretrained representations to Algerian hate speech detection. Second, the entire model is fine-tuned, allowing all Transformer parameters to be updated during training in order to better adapt the language representations to the target classification task.

Both Transformer models are fine-tuned under the same experimental protocol to ensure a fair comparison. Optimization is performed using the AdamW optimizer together with a linear warm-up schedule followed by linear learning-rate decay. The maximum input sequence length is fixed to 128 tokens with a batch size of 16. Hyperparameter selection and model validation follow the stratified five-fold cross-validation protocol described previously. 

\subsubsection{Proposed FAD-SA-GRU Architecture}
The previous approaches investigated in this work rely on a single semantic representation, either static word embeddings or contextual Transformer embeddings. Although these representations have demonstrated strong performance individually, they capture complementary linguistic information. Static embeddings, such as FastText and AraVec, effectively model lexical and morphological similarities, whereas contextual embeddings generated by Transformer models encode sentence-level semantics and long-range contextual dependencies. Given the linguistic complexity of Algerian Darija, characterized by code-switching between Arabic and French, the frequent use of Arabizi, and the absence of standardized orthography, relying on a single representation may not fully capture the semantic richness of social media comments.

To overcome these limitations, we propose \textbf{FAD-SA-GRU} (\textbf{F}usion of \textbf{A}lgerian \textbf{D}ialect Embeddings with \textbf{S}elf-\textbf{A}ttention \textbf{GRU}), a hybrid neural architecture that combines complementary embedding representations within a unified framework. The proposed model integrates trainable DZ FastText and DZ AraVec embeddings with contextual representations generated by a fine-tuned DziriBERT model. The resulting fused representation is projected into a common latent space, encoded by a Gated Recurrent Unit (GRU), and refined using a self-attention mechanism that automatically emphasizes the most informative tokens before the final binary classification stage.

\begin{figure}[]
    \centering
    \includegraphics[width=\linewidth]{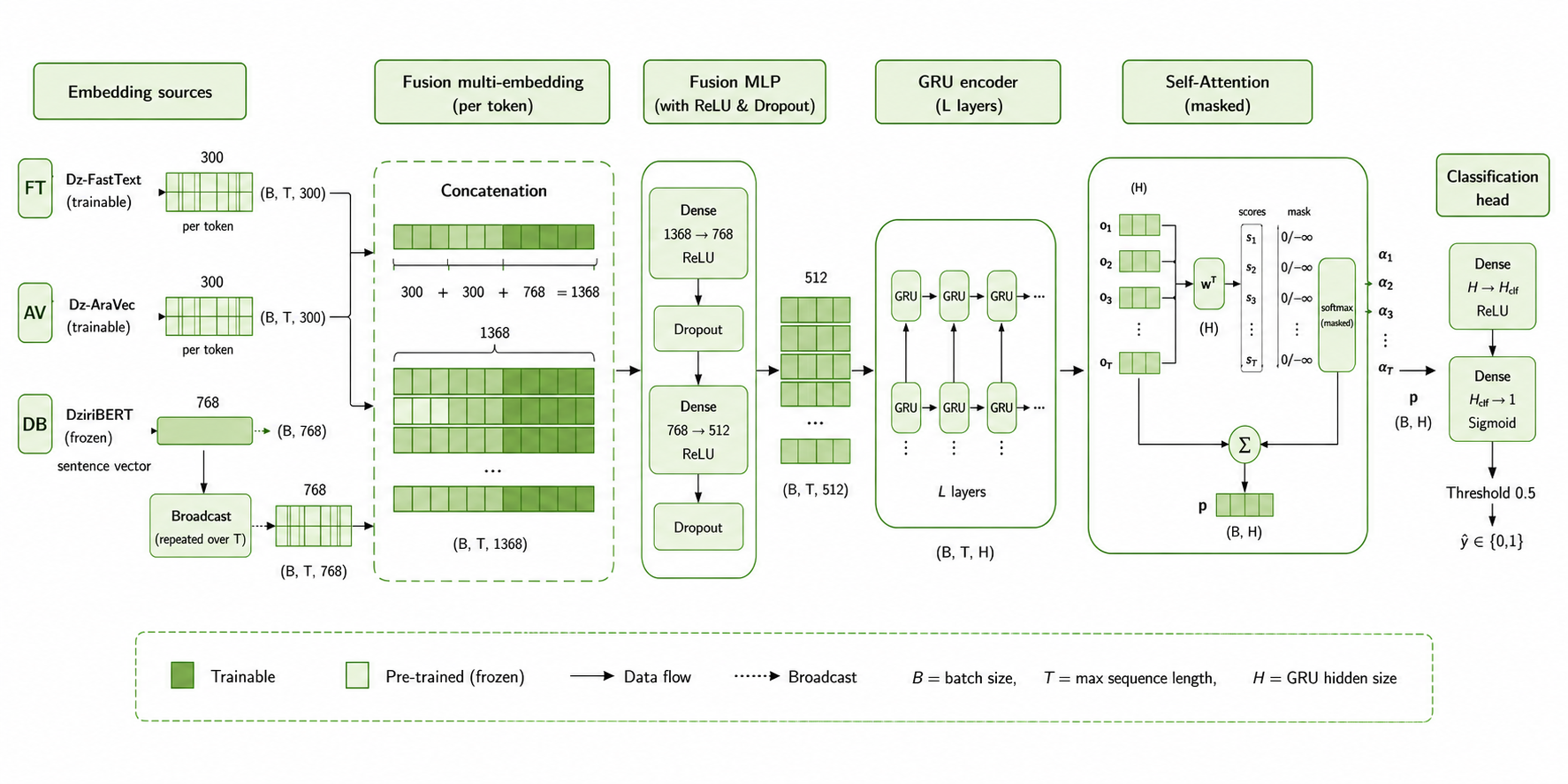}
    \caption{Architecture of the proposed FAD-SA-GRU model.}
    \label{fig:fad}
\end{figure}

Figure~\ref{fig:fad} illustrates the overall architecture of the proposed model.

\paragraph{Embedding Representations}
Three complementary embedding models are employed to represent each input comment.

\begin{itemize}

\item \textbf{DZ FastText.}
FastText provides subword-aware static word embeddings capable of handling spelling variations and out-of-vocabulary words commonly encountered in Algerian Darija. The pretrained 300-dimensional embeddings are further fine-tuned during training, producing a domain-adapted representation denoted by

\[
e_t^{FT}\in\mathbb{R}^{300}.
\]

\item \textbf{DZ AraVec.}
AraVec provides distributional semantic representations learned from large Arabic corpora. Similar to FastText, its pretrained 300-dimensional embeddings are fine-tuned on the target corpus, producing the representation

\[
e_t^{AV}\in\mathbb{R}^{300}.
\]

\item \textbf{Fine-tuned DziriBERT.}
Unlike the previous embeddings, DziriBERT generates contextual representations that depend on the complete input sentence. After fine-tuning on the hate speech classification task, the encoder is frozen and used to compute contextual sentence representations. Following masked mean pooling, the sentence embedding is obtained as

\[
e^{DB}=
\frac{\sum_{t=1}^{T}m_t h_t}
{\sum_{t=1}^{T}m_t}
\in\mathbb{R}^{768},
\]

where $h_t$ denotes the contextual representation of token $t$, $m_t\in\{0,1\}$ is the attention mask, and $T$ is the input sequence length. To reduce computational cost, these representations are precomputed once and cached before training the hybrid architecture.

\end{itemize}

\paragraph{Multi-Embedding Fusion}

Since DZ FastText and DZ AraVec generate token-level representations whereas DziriBERT produces a single contextual sentence embedding, the latter is broadcast across all token positions. For each token $t$, the three representations are concatenated to construct a unified feature vector

\[
f_t=
\left[
e_t^{FT}
\;\Vert\;
e_t^{AV}
\;\Vert\;
e^{DB}
\right]
\in\mathbb{R}^{1368},
\]

where $\Vert$ denotes vector concatenation.

Consequently, each token simultaneously benefits from complementary lexical, morphological, semantic, and contextual information. FastText contributes robust subword modeling, AraVec captures distributional semantic relationships, and DziriBERT provides sentence-level contextual knowledge. This complementary representation constitutes the main motivation behind the proposed hybrid architecture and serves as the input to the subsequent fusion network.

\paragraph{Fusion Network}

The concatenated representation $f_t$ has a dimensionality of 1368, which is relatively high for direct sequential modeling. To reduce the dimensionality while encouraging interactions between the three embedding sources, each fused vector is projected into a shared latent space using a multilayer perceptron (MLP). The projection is defined as

\[
g_t=\mathrm{MLP}(f_t)\in\mathbb{R}^{512},
\]

where the MLP consists of two fully connected layers following the architecture

\[
1368 \rightarrow 768 \rightarrow 512,
\]

with ReLU activation functions and dropout regularization applied between successive layers. The dropout rate is treated as a hyperparameter and optimized automatically during model selection. This projection produces a compact representation that preserves complementary semantic information while reducing redundancy before sequential encoding.

\paragraph{GRU Sequence Encoder}

The sequence of projected vectors

\[
(g_1,g_2,\ldots,g_T)
\]

is subsequently processed by a unidirectional Gated Recurrent Unit (GRU). Compared with standard recurrent neural networks, the GRU alleviates the vanishing gradient problem through reset and update gates while requiring fewer parameters than LSTM networks. This makes it particularly suitable for medium-sized datasets such as the corpus considered in this work.

The recurrent encoding is expressed as

\[
(o_1,\ldots,o_T)
=
\mathrm{GRU}(g_1,\ldots,g_T),
\qquad
o_t\in\mathbb{R}^{H},
\]

where $H$ denotes the hidden dimension of the recurrent layer.

Sequence packing is employed during training so that padding tokens are ignored by the recurrent computations, ensuring that only valid tokens contribute to the hidden representations. The hidden dimension and the number of stacked GRU layers are determined automatically through hyperparameter optimization.

\paragraph{Self-Attention Layer}

Although the GRU captures sequential dependencies, not every token contributes equally to the final classification decision. To enable the model to automatically identify the most informative words within a comment, a self-attention mechanism is applied to the GRU outputs.

For each hidden state $o_t$, an attention score is computed as

\[
e_t=w^{\top}o_t,
\]

where $w\in\mathbb{R}^{H}$ is a learnable attention vector.

The normalized attention coefficients are then obtained using the Softmax function

\[
\alpha_t
=
\frac{\exp(e_t)}
{\sum_{j=1}^{L_i}\exp(e_j)},
\]

where $L_i$ denotes the actual (non-padded) length of sample $i$. Padding positions are masked with $-\infty$ prior to the Softmax operation so that they receive zero attention weight.

Finally, a global context representation is computed as the weighted sum of all hidden states

\[
p
=
\sum_{t=1}^{L_i}
\alpha_t o_t
\in\mathbb{R}^{H}.
\]

Unlike average pooling or max pooling, the attention mechanism allows the network to dynamically focus on the words that are most discriminative for hate speech detection while assigning lower importance to less informative tokens.

\paragraph{Classification Layer}

The attention-pooled representation is passed to a fully connected classifier composed of one hidden layer followed by a sigmoid output neuron. The predicted probability is computed as

\[
\hat{y}
=
\sigma
\left(
W_{\mathrm{out}}
\,
\mathrm{ReLU}
\left(
W_{\mathrm{clf}}p+b_{\mathrm{clf}}
\right)
+b_{\mathrm{out}}
\right),
\]

where $W_{\mathrm{clf}}$ and $b_{\mathrm{clf}}$ denote the parameters of the hidden layer, $W_{\mathrm{out}}$ and $b_{\mathrm{out}}$ correspond to the output layer, and $\sigma(\cdot)$ represents the sigmoid activation function.

Binary predictions are obtained using a threshold of 0.5,

\[
\hat{c}
=
\mathbf{1}
\left[
\hat{y}\ge0.5
\right],
\]

where $\mathbf{1}[\cdot]$ denotes the indicator function.

The network is optimized using the Binary Cross-Entropy (BCE) loss

\[
\mathcal{L}
=
-
\frac{1}{N}
\sum_{i=1}^{N}
\left[
y_i\log(\hat{y}_i)
+
(1-y_i)\log(1-\hat{y}_i)
\right],
\]

where $N$ is the batch size, $y_i$ is the ground-truth label, and $\hat{y}_i$ is the predicted probability for sample $i$.

The~\autoref{tab:fad-dims} summarizes the dimensions at each stage of the pipeline.

\begin{table}[h]
\centering
\caption{Dimensions of the representations at each stage of FAD-SA-GRU.}
\label{tab:fad-dims}
\begin{tabular}{lll}
\toprule
Stage & Component & Output dimension \\
\midrule
Dz-FastText & Trainable embedding & $(B, T, 300)$ \\
Dz-AraVec & Trainable embedding & $(B, T, 300)$ \\
DziriBERT (broadcast) & Repeated sentence vector & $(B, T, 768)$ \\
Concatenation & Fusion of the three sources & $(B, T, 1368)$ \\
Fusion MLP & $1368 \to 768 \to 512$ & $(B, T, 512)$ \\
GRU & Hidden dimension $H \in [64, 256]$ & $(B, T, H)$ \\
Self-Attention & Weighting and aggregation of real tokens & $(B, H)$ \\
Classifier & $H \to H_{clf} \to 1$ & $(B, 1)$ \\
\bottomrule
\end{tabular}
\end{table}

Overall, the proposed FAD-SA-GRU architecture combines complementary static and contextual embeddings, nonlinear feature fusion, sequential modeling through GRU, and adaptive token weighting via self-attention within a single end-to-end framework. This design enables the model to simultaneously exploit lexical, semantic, and contextual information, making it particularly well suited to the linguistic variability of Algerian Darija.

The optimal hyperparameters of the FAD-SA-GRU approach are: hidden dimension 96, 2 GRU layers, MLP dropout 0.1, classifier dropout
0.1, classifier hidden size 256, learning rate $5.13 \times 10^{-5}$, weight decay 0.00428,
batch size 128, gradient clipping 0.59, and 10 training epochs. \autoref{tab:fad-hyperparams} summarizes these settings.

\begin{table}[h!]
\centering
\caption{Hyperparameters of the FAD-SA-GRU model.}
\label{tab:fad-hyperparams}
\resizebox{\textwidth}{!}{%
\begin{tabular}{|c|c|c|c|c|c|c|c|c|}
\hline
\multirow{4}{*}{\textbf{Model}} & \multicolumn{8}{c|}{\textbf{Hyperparameters}} \\
\cline{2-9}
& \texttt{hidden\_dim} & \texttt{n\_layers} & \texttt{dropout\_pre} & \texttt{dropout} & \texttt{dropout\_post} & \texttt{mlp\_dropout} & \texttt{clf\_dropout} & \texttt{clf\_hidden} \\
\cline{1-1}
\multirow{-3}{*}{FAD-SA-GRU} & 96 & 2 & 0.5 & 0.2 & 0.5 & 0.1 & 0.1 & 256 \\
\cline{2-9}
& \texttt{learning\_rate} & \texttt{weight\_decay} & \texttt{batch\_size} & \texttt{clip\_norm} & \texttt{best\_factor} & \texttt{best\_patience} & \texttt{optimal\_epoch} & \\
\cline{2-9}
& 5.13E-0.5 & 0.00428 & 128 & 0.59 & 0.4 & 0 & 10 & X \\
\hline
\end{tabular}%
}
\end{table}

\section{Experiments and Results}
\label{sec:results}

\subsection{Learning Curves}
Learning curves obtained during stratified five-fold cross-validation were analyzed to assess the convergence behavior and generalization ability of each family of models.

\begin{figure}[h!]
    \centering
    \includegraphics[width=\linewidth]{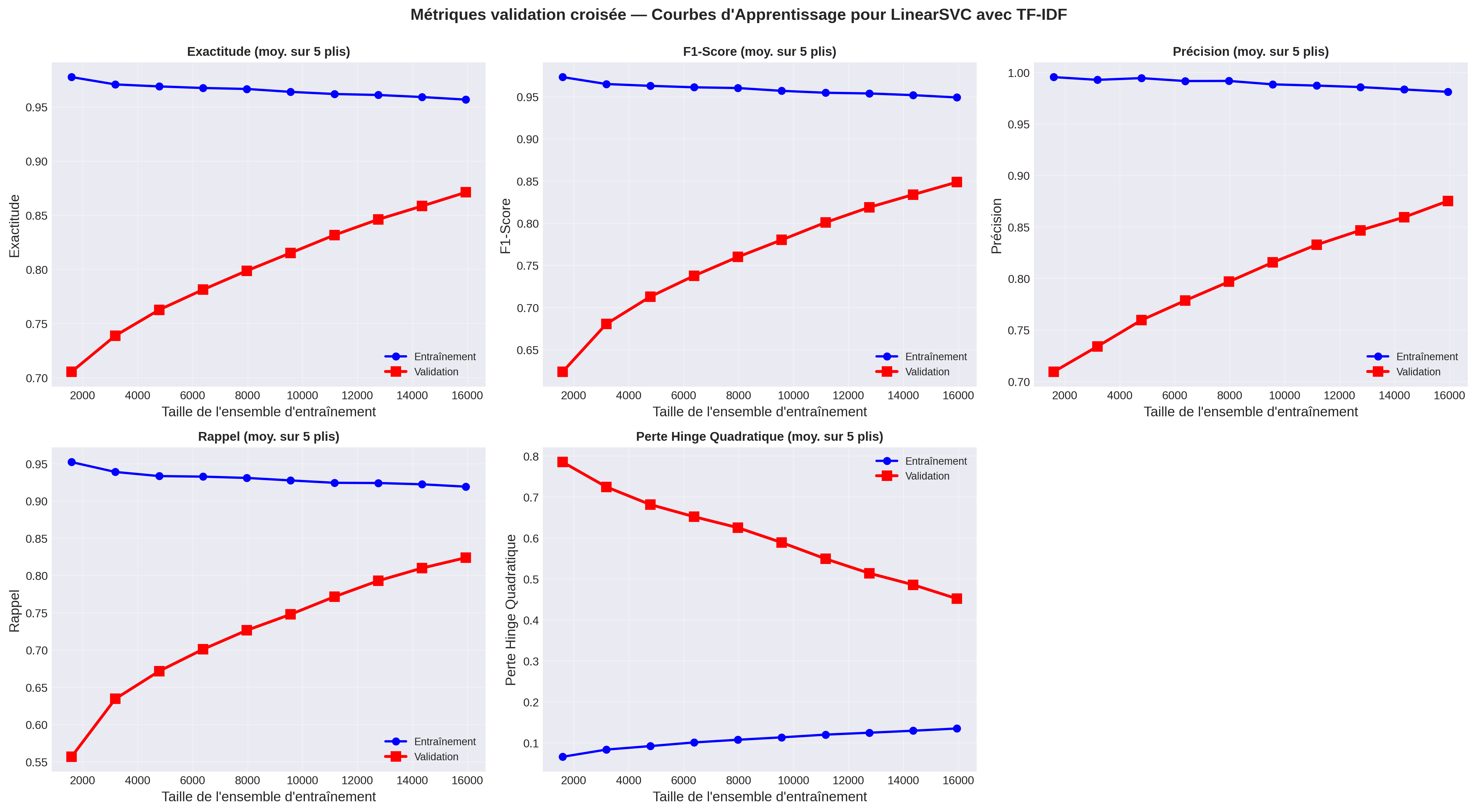}
    \caption{Learning curves of the TF--IDF + LinearSVC model.}
    \label{fig:svc-curve}
\end{figure}

Figure~\ref{fig:svc-curve} presents the learning curves of the TF--IDF + LinearSVC classifier. As the training set gradually increases, the validation performance consistently improves while the gap between the training and validation curves decreases, indicating good generalization and limited overfitting. The validation Squared Hinge Loss decreases steadily, confirming the stability of the learned decision boundary.

\begin{figure}[h!]
    \centering
    \includegraphics[width=0.9\linewidth]{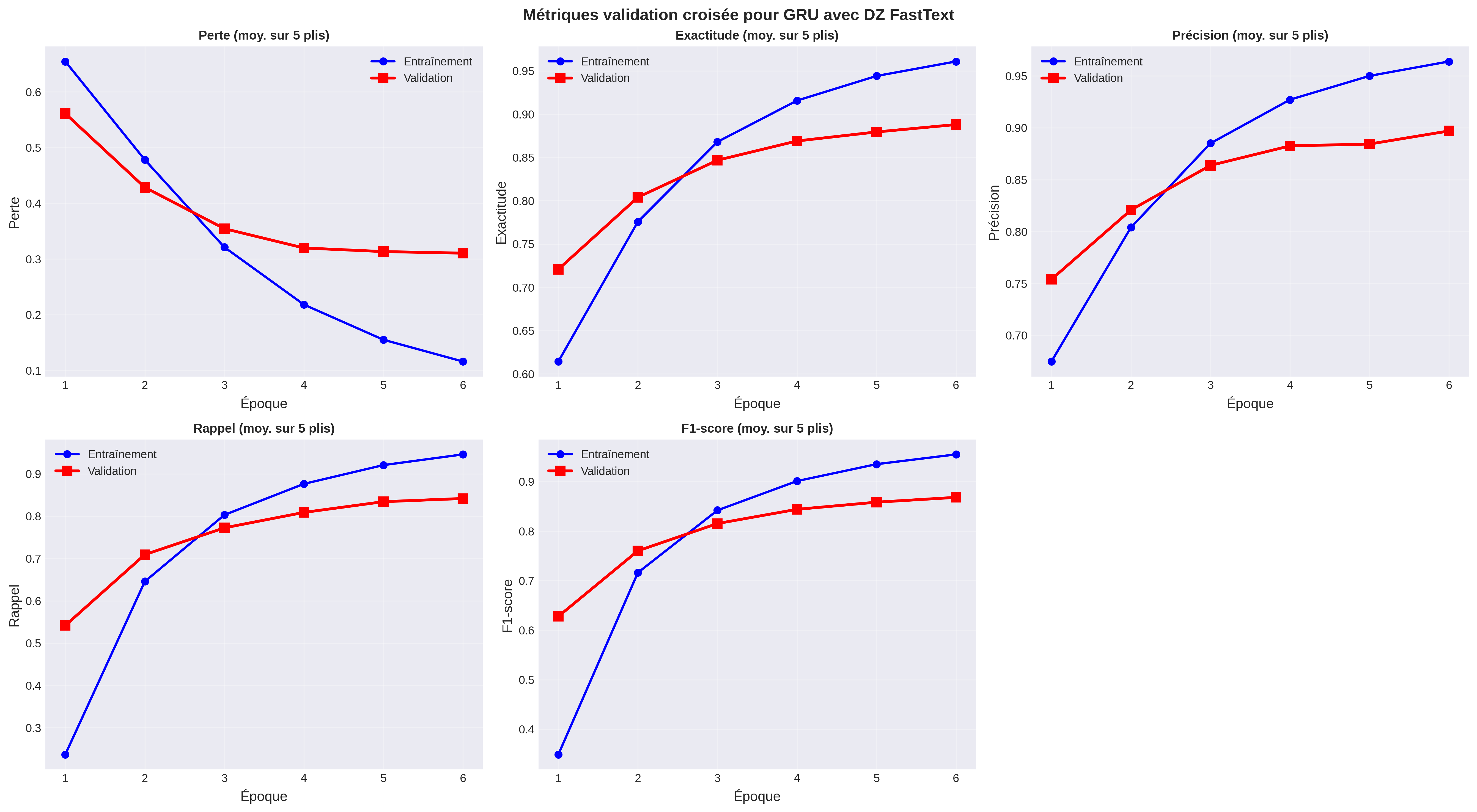}
    \caption{Learning curves of the Dz FastText + GRU model after early stopping.}
    \label{fig:gru-nooverfit}
\end{figure}

For the recurrent models, the Dz FastText + GRU architecture exhibited the most stable optimization process. As shown in Figure~\ref{fig:gru-nooverfit}, validation performance improved during the first epochs before stabilizing. Cross-validation identified epoch 6 as the optimal stopping point, after which validation loss started to increase, indicating the onset of overfitting. The final model was therefore trained for six epochs.

\begin{figure}[h!]
    \centering
    \includegraphics[width=0.9\linewidth]{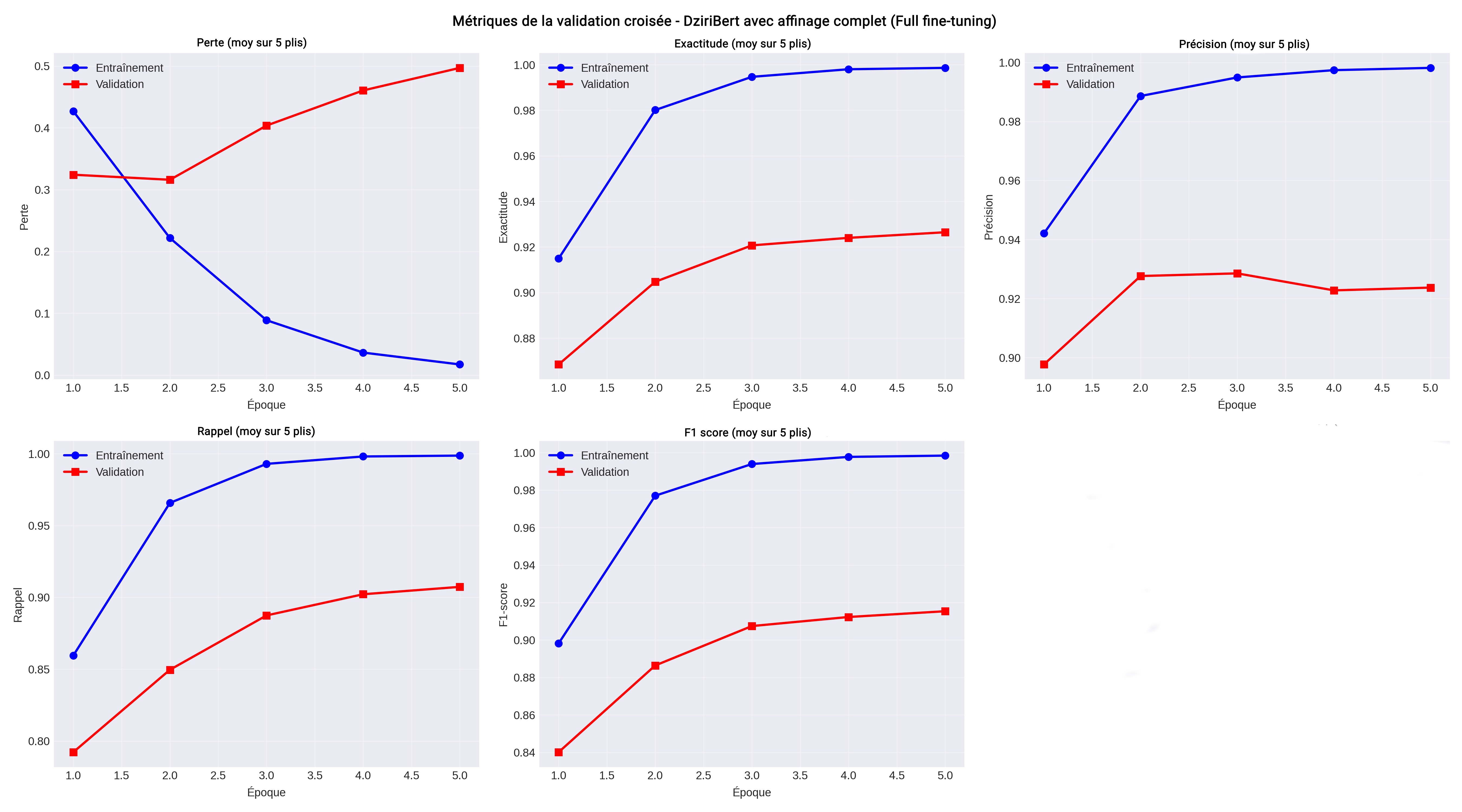}
    \caption{Learning curves of DziriBERT before early stopping.}
    \label{fig:dziribert-overfit}
\end{figure}

Figure~\ref{fig:dziribert-overfit} illustrates the training dynamics of the fully fine-tuned DziriBERT model. Validation loss reaches its minimum after the second epoch and subsequently increases, revealing overfitting despite continued improvement on the training set.

\begin{figure}[h!]
    \centering
    \includegraphics[width=0.9\linewidth]{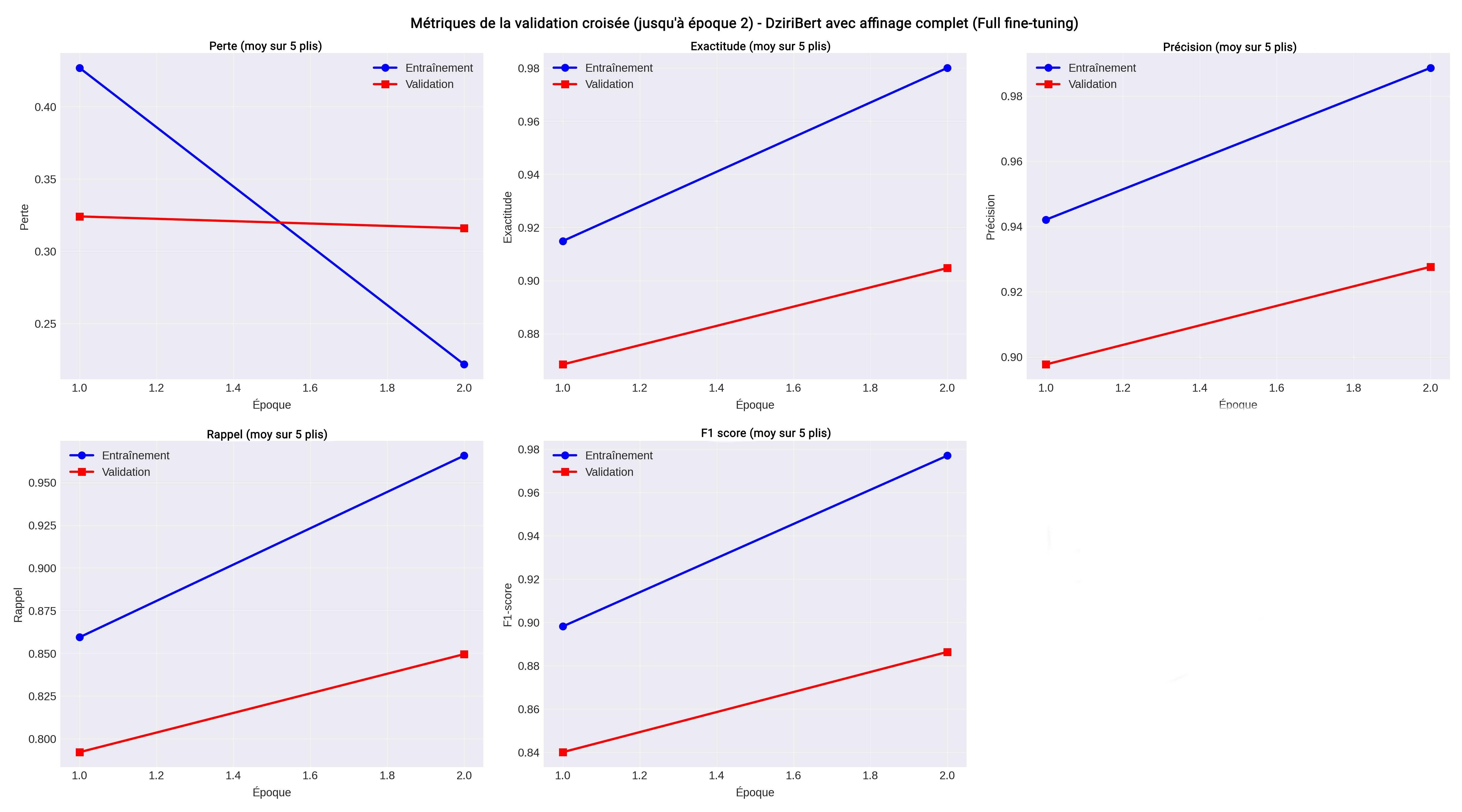}
    \caption{Learning curves of the final DziriBERT model.}
    \label{fig:dziribert-nooverfit}
\end{figure}

Consequently, the final DziriBERT model was trained for two epochs only, yielding the learning curves shown in Figure~\ref{fig:dziribert-nooverfit}. This configuration consistently outperformed the multilingual BERT baseline.

\begin{figure}[h!]
    \centering
    \includegraphics[width=\linewidth]{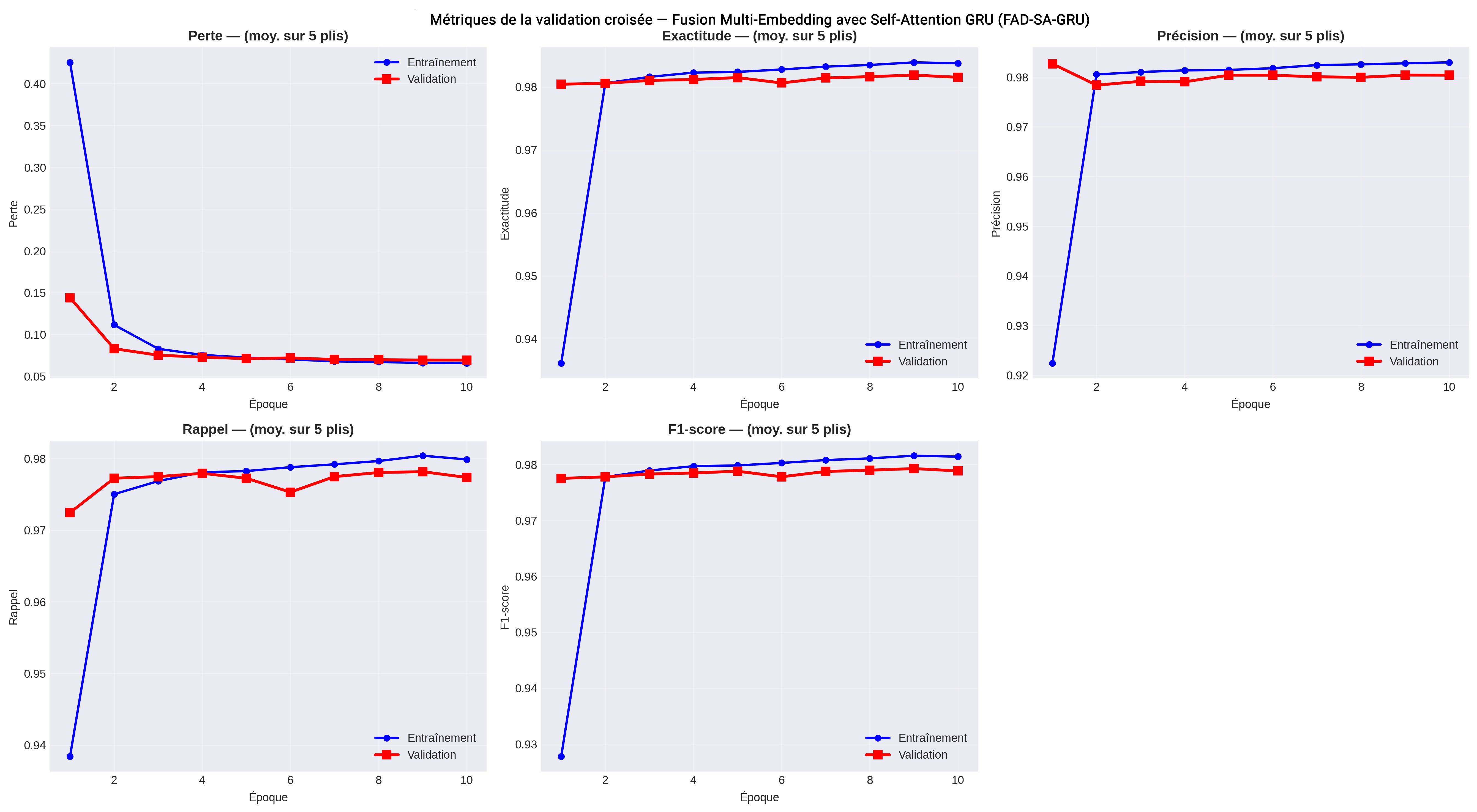}
    \caption{Learning curves of the proposed FAD-SA-GRU model.}
    \label{fig:fad-curve}
\end{figure}

The proposed FAD-SA-GRU model exhibits the most stable optimization behavior among all evaluated architectures. As illustrated in Figure~\ref{fig:fad-curve}, both training and validation metrics improve consistently throughout training while remaining closely aligned, indicating excellent generalization and no observable overfitting. Consequently, the final model was trained for the full ten epochs.

\subsection{Evaluation Metrics}
Model performance was evaluated on the independent test set using five standard binary classification metrics: Accuracy, Precision, Recall, F1-score, and ROC-AUC.

Table~\ref{tab:all-metrics} summarizes the performance of all evaluated models.

\begin{table}[h]
\centering
\caption{Metrics of all evaluated models on the test set.}
\label{tab:all-metrics}

\resizebox{\textwidth}{!}{%
\begin{tabular}{llccccc}
\toprule
Family & Model & Accuracy & Precision & Recall & F1-score & ROC-AUC \\
\midrule
\multirow{3}{*}{Machine Learning}
 & TF-IDF + LinearSVC & 0.886 & 0.909 & 0.822 & 0.863 & 0.949 \\
 & TF-IDF + Logistic Regression & 0.878 & 0.907 & 0.805 & 0.853 & 0.938 \\
 & TF-IDF + SGDClassifier & 0.878 & 0.905 & 0.806 & 0.852 & 0.939 \\
\midrule
\multirow{9}{*}{Deep Learning}
 & FastText + RNN & 0.757 & 0.719 & 0.733 & 0.725 & 0.844 \\
 & AraVec + RNN & 0.684 & 0.711 & 0.469 & 0.565 & 0.721 \\
 & Dz FastText + RNN & 0.917 & 0.932 & 0.875 & 0.902 & 0.970 \\
 & FastText + LSTM & 0.771 & 0.745 & 0.726 & 0.735 & 0.862 \\
 & AraVec + LSTM & 0.659 & 0.655 & 0.465 & 0.544 & 0.697 \\
 & Dz FastText + LSTM & 0.925 & 0.924 & 0.903 & 0.914 & 0.975 \\
 & FastText + GRU & 0.781 & 0.816 & 0.646 & 0.721 & 0.872 \\
 & AraVec + GRU & 0.682 & 0.705 & 0.471 & 0.565 & 0.730 \\
 & Dz FastText + GRU & 0.928 & 0.927 & 0.905 & 0.916 & 0.976 \\
\midrule
\multirow{4}{*}{Pre-trained Transformers}
 & DziriBERT without fine-tuning & 0.414 & 0.383 & 0.556 & 0.454 & 0.398 \\
 & DziriBERT with full fine-tuning & 0.929 & 0.937 & 0.900 & 0.918 & 0.976 \\
 & BERT-Base-Multilingual-Cased without fine-tuning & 0.493 & 0.383 & 0.260 & 0.310 & 0.445 \\
 & BERT-Base-Multilingual-Cased with full fine-tuning & 0.879 & 0.888 & 0.829 & 0.857 & 0.945 \\
\midrule
Our hybrid approach & \textbf{FAD-SA-GRU} & \textbf{0.932} & \textbf{0.934} & \textbf{0.910} & \textbf{0.921} & \textbf{0.970} \\
\bottomrule
\end{tabular}%
}
\end{table}

Several observations can be drawn from these results.

First, among the classical machine learning approaches, LinearSVC achieved the best overall performance, obtaining an F1-score of 86.3\% and a ROC-AUC of 94.9\%. This confirms the effectiveness of linear classifiers combined with TF--IDF representations for sparse textual data.

Second, fine-tuning substantially improved the performance of recurrent neural networks. While the pretrained FastText and AraVec embeddings produced relatively modest results, adapting FastText to the target corpus (Dz FastText) increased the F1-score by nearly 20 percentage points. Among the recurrent architectures, GRU achieved the highest performance (F1 = 91.6\%), closely followed by LSTM.

Third, the Transformer-based models clearly demonstrate the importance of task-specific fine-tuning. Without fine-tuning, both DziriBERT and mBERT performed close to random guessing. After full fine-tuning, DziriBERT achieved an F1-score of 91.8\%, substantially outperforming multilingual BERT, highlighting the benefit of a language model pretrained specifically on Algerian Darija.

Finally, the proposed \textbf{FAD-SA-GRU} architecture achieved the best performance across nearly all evaluation metrics, reaching an Accuracy of 93.2\%, Precision of 93.4\%, Recall of 91.0\%, and an F1-score of 92.1\%. These results demonstrate that combining complementary static and contextual embeddings through a fusion strategy, followed by GRU sequence modeling and self-attention, provides a more discriminative representation than any individual embedding model considered in this study.

\section{Analysis and Discussion}
\label{sec:discussion}
The experimental results provide several insights into hate speech detection in Algerian Darija. Beyond the numerical performance of individual models, they highlight the importance of domain-specific representations, contextual modeling, and embedding complementarity when dealing with a linguistically heterogeneous dialect.

\subsection{Comparative Analysis of Model Families}
The comparison between the four investigated approaches reveals a progressive improvement in performance as richer linguistic representations are introduced. Classical machine learning methods constitute a strong baseline, particularly LinearSVC combined with TF--IDF features. Their competitive performance confirms that lexical information alone is sufficient to detect a large proportion of explicit hate speech, where offensive vocabulary is highly discriminative. However, these models rely on sparse bag-of-words representations and therefore ignore word order, contextual information, and semantic relationships between terms. Consequently, they struggle to recognize implicit hate speech expressed through paraphrases, idiomatic expressions, or contextual cues.

The recurrent neural networks overcome part of these limitations by modeling sequential dependencies between words. A particularly important observation is the impact of embedding adaptation. While the original pretrained FastText and AraVec embeddings yielded relatively modest performance, fine-tuning FastText on the target corpus produced substantial improvements for all recurrent architectures. This result suggests that general-purpose embeddings do not adequately represent the lexical variations and code-switching patterns that characterize Algerian social media, whereas domain-adapted embeddings are able to capture these linguistic specificities more effectively.

Among the recurrent architectures, GRU consistently outperformed both the standard RNN and LSTM. Although LSTM introduces an additional memory cell intended to capture long-term dependencies, the comments contained in the corpus are generally short, limiting the practical advantage of this additional complexity. GRU therefore offers a better trade-off between model capacity, computational efficiency, and generalization.

The Transformer-based models further demonstrate the importance of contextual language representations. Without fine-tuning, both DziriBERT and mBERT produced performances close to random guessing, indicating that pretrained language representations alone cannot be transferred directly to the hate speech classification task. After fine-tuning, however, DziriBERT achieved performance comparable to the best recurrent model and clearly outperformed multilingual BERT. This behavior confirms that language models pretrained on Algerian dialect data capture lexical, syntactic, and semantic patterns that cannot be learned efficiently by multilingual models trained on more general corpora.

The proposed FAD-SA-GRU architecture achieved the highest overall performance. Its improvement over the individual models demonstrates that the three embedding sources provide complementary information rather than redundant representations. DZ FastText contributes robust subword representations that are resilient to spelling variations and Arabizi writing, DZ AraVec captures broader semantic relationships learned from Arabic corpora, while DziriBERT provides contextual sentence-level information. The fusion module enables the model to jointly exploit these heterogeneous representations before sequential encoding, while the self-attention mechanism further improves discrimination by assigning higher importance to the most informative tokens. The consistent improvement across all evaluation metrics indicates that combining complementary representations is more effective than relying on a single embedding model.

\subsection{Discussion}
The obtained results are consistent with recent advances in hate speech detection reported for both high-resource languages and Algerian Darija. Previous studies have shown that contextual language models generally outperform traditional machine learning approaches, while hybrid neural architectures often provide additional gains by combining multiple sources of linguistic information. The results of this work confirm these observations in the specific context of Algerian Darija.

An important finding concerns the role of domain adaptation. The performance gap between pretrained embeddings and their fine-tuned counterparts highlights the dynamic nature of social media language. Algerian Darija evolves rapidly, with frequent lexical innovations, spelling variations, code-switching, and Arabizi usage. Models trained on generic corpora are therefore unable to fully capture these characteristics without task-specific adaptation.

Despite the strong overall performance, several challenges remain. Most classification errors correspond to implicit hate speech, sarcasm, irony, or culturally dependent expressions whose interpretation requires pragmatic knowledge beyond the textual content itself. Similarly, comments containing political figures, ethnic references, or community names may appear in both hateful and neutral contexts, making lexical information alone insufficient for reliable classification. These observations suggest that future systems should incorporate additional contextual information, conversational history, or external knowledge sources to better model implicit forms of hate speech.

Overall, the experimental results demonstrate that effective hate speech detection in Algerian Darija requires the combination of domain-specific embeddings, contextual language models, and adaptive attention mechanisms. The proposed FAD-SA-GRU architecture successfully integrates these complementary components, providing a robust solution for the automatic moderation of Algerian social media content while remaining sufficiently general to be extended to other Maghrebi dialects.

\section{Conclusion}
\label{sec:conclusion}
This paper presented a comprehensive framework for the automatic detection of hate speech in Algerian Arabic (Darija) using social media comments collected from the official Facebook page of ATM Mobilis. Owing to the linguistic complexity of Darija, characterized by code-switching between Arabic and French, the widespread use of Arabizi, and the absence of standardized orthography, several complementary approaches were investigated to identify the most effective strategy for hate speech classification.

A comparative evaluation was conducted across four families of methods, namely traditional machine learning algorithms, recurrent deep learning architectures, pretrained Transformer models, and a novel hybrid architecture, \textbf{FAD-SA-GRU}. By combining DZ FastText, DZ AraVec, and DziriBERT embeddings through a multi-embedding fusion strategy enhanced with a self-attention mechanism, the proposed model achieved the best overall performance, obtaining an accuracy of 93.2\%, a precision of 93.4\%, a recall of 91.0\%, an F1-score of 92.1\%, and a ROC-AUC of 97.0\%. These results demonstrate that integrating complementary static and contextual representations significantly improves the detection of hate speech in low-resource dialectal Arabic. In addition, the development of a RESTful API integrating all evaluated models provides a practical solution for deploying automated hate speech moderation systems.

Despite these promising results, several challenges remain. The size of the annotated corpus is still relatively limited, while the dynamic nature of social media continuously introduces new vocabulary, spelling variations, and linguistic expressions. Moreover, implicit hate speech, sarcasm, irony, and culturally dependent expressions remain difficult to identify accurately and continue to represent important open research challenges.

Future work will focus on expanding the corpus with more diverse data sources, investigating continual learning strategies to adapt to the evolution of online language, and extending the proposed framework to multimodal hate speech detection by incorporating textual and visual information. Additional research will also explore explainable artificial intelligence techniques, such as SHAP, LIME, and attention visualization, to improve model transparency and interpretability. 
Finally, the proposed FAD-SA-GRU architecture could be adapted to other Maghrebi dialects, including Tunisian and Moroccan Arabic, as well as Amazigh varieties spoken in Algeria, thereby contributing to the development of robust NLP resources for underrepresented languages and dialects.

\bibliographystyle{unsrtnat}
\bibliography{references} 

@String{Computing = "Computing" }

@String{Computer = "{IEEE} Computer" }

@article{lanasri2023hate,
  title={Hate speech detection in algerian dialect using deep learning},
  author={Lanasri, Dihia and Olano, Juan and Klioui, Sifal and Lee, Sin Liang and Sekkai, Lamia},
  journal={arXiv preprint arXiv:2309.11611},
  year={2023}
}

@article{khezzar2023,
  title={arhatedetector: detection of hate speech from standard and dialectal arabic tweets},
  author={Khezzar, Ramzi and Moursi, Abdelrahman and Al Aghbari, Zaher},
  journal={Discover Internet of Things},
  year={2023},
  doi={10.1007/s43926-023-00030-9}
}

@article{saleh2023,
  title={Detection of hate speech using bert and hate speech word embedding with deep model},
  author={Saleh, Hind and Alhothali, Areej and Moria, Kawthar},
  journal={Applied Artificial Intelligence},
  volume={37},
  number={1},
  pages={e2166719},
  year={2023},
  doi={10.1080/08839514.2023.2166719}
}

@article{chared2025,
  title={AI-driven detection of hate speech on social media: a case study in the french language},
  author={Chared, Zahim and Jantet, Clément and Ravix, Calliste and Salmi, Robin and Hashmi, Ehtesham and Yayilgan, Sule Yildirim},
  journal={Cluster Computing},
  volume={28},
  number={4},
  pages={811},
  year={2025},
  doi={10.1007/s10586-025-05553-0}
}

@article{bouchal2023,
  title={Arabic hate speech and social networks offensive language detection},
  author={Bouchal, Hakim and Belaid, Ahror},
  journal={Information Processing at the Digital Age Journal},
  year={2023},
  url={https://asjp.cerist.dz/en/article/240580}
}

@article{kumar2024,
  title={A hybrid deep bilstm-cnn for hate speech detection in multi-social media},
  author={Kumar, Ashwini and Kumar, Santosh and Passi, Kalpdrum and Mahanti, Aniket},
  journal={ACM Trans. Asian Low-Resour. Lang. Inf. Process.},
  volume={23},
  number={8},
  year={2024},
  doi={10.1145/3657635}
}

@article{delaval2026,
  title={Toxifrench: Benchmarking and enhancing language models via cot fine-tuning for french toxicity detection},
  author={Delaval, Axel and Yang, Shujian and Wang, Haicheng and Qiu, Han and Lu, Jialiang},
  journal={arXiv preprint},
  year={2026},
  doi={10.48550/arXiv.2508.11281}
}

@misc{guellil2021,
  title={Sexism detection: The first corpus in algerian dialect with a code-switching in arabic/french and english},
  author={Guellil, Imane and Adeel, Ahsan and Azouaou, Faical and Boubred, Mohamed and Houichi, Yousra and Moumna, Akram Abdelhaq},
  year={2021},
  url={https://arxiv.org/pdf/2104.01443}
}

@article{boucherit2022,
  title={Offensive language detection in under-resourced algerian dialectal arabic language},
  author={Boucherit, Oussama and Abainia, Kheireddine},
  journal={ACM Computing Surveys},
  year={2022},
  doi={10.48550/arXiv.2203.10024}
}

@article{mazari2023,
  title={Deep learning-based analysis of algerian dialect dataset targeted hate speech, offensive language and cyberbullying},
  author={Mazari, Ahmed Cherif and Kheddar, Hamza},
  journal={International Journal of Computing and Digital System},
  year={2023},
  doi={10.12785/ijcds/130177}
}

@article{abdedaiem2024,
  title={Fassila: A corpus for algerian dialect fake news detection and sentiment analysis},
  author={Abded{\"a}iem, Amin and Dahou, Abdelhalim Hafedh and Cheragui, Mohamed Amine and Mathiak, Brigitte},
  journal={Procedia Computer Science},
  year={2024}
}

@article{perez2023assessing,
  title={Assessing the impact of contextual information in hate speech detection},
  author={P{\'e}rez, Juan Manuel and others},
  journal={IEEE Access},
  volume={11},
  pages={28354--28365},
  year={2023},
  doi={10.1109/ACCESS.2023.3258973}
}

@article{abdaoui2021dziribert,
  title={Dziribert: a pre-trained language model for the algerian dialect},
  author={Abdaoui, Amine and Berrimi, Mohamed and Oussalah, Mourad and Moussaoui, Abdelouahab},
  journal={arXiv preprint arXiv:2109.12346},
  year={2021},
  doi={10.48550/arXiv.2109.12346}
}
\end{document}